\pdfoutput=1

\documentclass[11pt]{article}

\usepackage[]{latex/acl}

\usepackage{times}
\usepackage{multirow}
\usepackage{latexsym}
\usepackage{soul}
\usepackage{comment}

\usepackage[T1]{fontenc}

\usepackage[utf8]{inputenc}

\usepackage{microtype}

\usepackage{inconsolata}

\usepackage{amsmath,amsfonts,bm}

\def\eqref#1{equation~\ref{#1}}

\def\1{\bm{1}}

\def\vx{{\bm{x}}}
\def\vy{{\bm{y}}}

\DeclareMathAlphabet{\mathsfit}{\encodingdefault}{\sfdefault}{m}{sl}
\SetMathAlphabet{\mathsfit}{bold}{\encodingdefault}{\sfdefault}{bx}{n}

\DeclareMathOperator*{\argmin}{arg\,min}

\usepackage{graphicx}
\usepackage{diagbox}
\usepackage{adjustbox}
\usepackage{pgfplots}
\usepackage{subfigure}
\usepackage{booktabs} 

\usepackage{url}

\usepackage{breakurl}
\usepackage[breaklinks]{hyperref}

\definecolor{darkblue}{rgb}{0.0,0.0,0.65}

\DeclareUnicodeCharacter{2060}{\nolinebreak}

\hypersetup{
  colorlinks = true,
  citecolor  = darkblue,
  linkcolor  = darkblue,
  citecolor  = darkblue,
  filecolor  = darkblue,
  urlcolor   = darkblue,
}

\usepackage{subfiles}
\usepackage{url}
\usepackage{caption}
\usepackage{graphicx}
\usepackage{amsfonts}
\usepackage{amsmath}
\usepackage{amsthm}
\usepackage{subfigure}
\usepackage{amssymb}
\usepackage{mathtools}
\usepackage{bbm}
\usepackage{algorithm}
\usepackage{algpseudocode}
\usepackage{enumitem}
\usepackage{amssymb}
\usepackage{bbm}
\usepackage{wrapfig}

\theoremstyle{definition}

\newcommand{\Sref}[1]{\S\ref{#1}}

\usepackage{xcolor}

\newcommand{\LM}{P_{\text{LM}}}
\newcommand{\embd}{M_{\text{embd}}}
\newcommand{\method}{\textsc{SemStamp}}
\newcommand{\shortmethod}{\textsc{SStamp}}

\newcommand{\secvsabove}{\vspace{-1mm}}
\newcommand{\secvsbelow}{\vspace{-1mm}}
\newcommand{\subsecvs}{\vspace{-1mm}}

\newcommand{\figvsbottom}{\vspace{-3mm}}
\newcommand{\paravs}{\vspace{-1.5mm}}
\newcommand{\eqvs}{\vspace{-1.5mm}}

\newlength\myindent
\title{\method: A Paraphrase-Robust  Watermark}
\title{\method: A Watermark Robust to Semantic Paraphrases}
\title{Optimizing Paraphrasing Robustness for Natural Language Watermark}
\title{A Semantic Watermark Algorithm for Language Model Generation with Robustness to Paraphrase Attacks}
\title{\method: Paraphrase-Robust Semantic Watermark}
\title{\method: A Paraphrase-Robust Semantic Watermark}
\title{\method: Semantic Watermarking with Paraphrase Invariance}
\title{\method: A Semantic Watermark}
\title{\method: Semantically Watermarking Text Generation with Paraphrastic Robustness}
\title{\method: Semantic Watermarking \\ with  Paraphrastic Robustness}
\title{\method: Paraphrase-Robust Watermarked \\ 
Text Generation}
\title{\method: Semantically Watermarking Text via Paraphrastic Robustness}
\title{\method: Semantically Watermarking Language Model Texts with Paraphrastic Robustness}
\title{\method: Semantic Watermarking Text Generation with Paraphrastic Robustness}
\title{\method: Semantic Watermarked Generation \\ with Paraphrastic Robustness}
\title{\vspace*{-0.5in}
{{\small \hfill NAACL'24}\\
\vspace*{.25in}} \method: A Semantic Watermark with Paraphrastic Robustness for Text Generation}

\author{Abe Bohan Hou\textsuperscript{$\clubsuit$}\thanks{Equal Contribution. All three are corresponding authors.} \quad Jingyu Zhang\textsuperscript{$\clubsuit$}\footnotemark[1] \quad Tianxing He\textsuperscript{$\heartsuit$}\footnotemark[1] \\ \bf{Yichen Wang\textsuperscript{$\diamondsuit$} \quad Yung-Sung Chuang\textsuperscript{$\spadesuit$} \quad Hongwei Wang\textsuperscript{$\ddagger$} \quad Lingfeng Shen\textsuperscript{$\clubsuit$}} \\ \bf{Benjamin Van Durme\textsuperscript{$\clubsuit$} \quad Daniel Khashabi\textsuperscript{$\clubsuit$} \quad Yulia Tsvetkov\textsuperscript{$\heartsuit$}}\\
\textsuperscript{$\clubsuit$}Johns Hopkins University \quad \textsuperscript{$\heartsuit$}University of Washington \quad \textsuperscript{$\diamondsuit$}{Xi'an Jiaotong University}
\\
\textsuperscript{$\spadesuit$}{Massachusetts Institute of Technology} \quad 
\textsuperscript{$\ddagger$}{Tencent AI Lab}\\
\texttt{\{bhou4, jzhan237\}@jhu.edu\quad goosehe@cs.washington.edu}
}

\begin{document}

\maketitle

\begin{abstract}
Existing watermarked generation algorithms employ \emph{token-level} designs and therefore, are vulnerable to paraphrase attacks. 
To address this issue, we introduce watermarking on the \emph{semantic representation} of sentences.
We propose \method, a robust sentence-level semantic watermarking algorithm that uses locality-sensitive hashing (LSH) to partition the semantic space of sentences. The algorithm encodes and LSH-hashes a candidate sentence generated by a language model, and conducts rejection sampling until the sampled sentence falls in watermarked partitions in the semantic embedding space. 
To test the paraphrastic robustness of watermarking algorithms, we propose a ``bigram paraphrase'' attack that produces paraphrases with small bigram overlap with the original sentence. 
This attack is shown to be effective against existing token-level watermark algorithms, while posing only minor degradations to \method.
Experimental results show that our novel semantic watermark algorithm is not only more robust than the previous state-of-the-art method on various paraphrasers and domains, but also better at preserving the quality of generation.\footnote{We release code in \url{https://github.com/bohanhou14/SemStamp}}
\end{abstract}
\secvsabove
\section{Introduction}
\secvsbelow
\label{sec:intro}

{This work focuses on algorithms for detecting machine-generated text via 
 \textit{watermarked generation}---adding signatures during text generation which are algorithmically detectable, yet are imperceptible to human eye~\citep{10.1007/3-540-45496-9_14}. 
This problem is of extreme importance now that large language models (LLMs) such as GPT-4 \citep{openai2023gpt4}  generate realistic text, increasing risks of LLM misuse, such as generation of misinformation, impersonation, and copyright infringements \citep{weidinger2021ethical,ippolito2022preventing,pagnoni-etal-2022-threat, watermark-gov}. 
}



The dominant body of recent works on watermarked generation operate by injecting token-level signatures during decoding time~\citep[\textit{i.a.}]{Kuditipudi2023RobustDW, Yoo2023RobustMN, Wang2023TowardsCT, Christ2023UndetectableWF, Fu2023WatermarkingCT}.
As a representative example, 
\citet{kirchenbauer2023watermark} 
propose a watermarked generation algorithm that injects watermark signals that are extracted based on the previously generated \emph{tokens}.  
Despite its efficiency, follow-up work has shown that corrupting the generated text, especially paraphrasing, could weaken its robustness \citep{krishna2023paraphrasing, sadasivan2023aigenerated, kirchenbauer2023reliability}. 

\begin{figure*}[ht]
    \centering
    \vspace{-4mm}
    \hspace{-4mm}\includegraphics[width=1.02\textwidth]{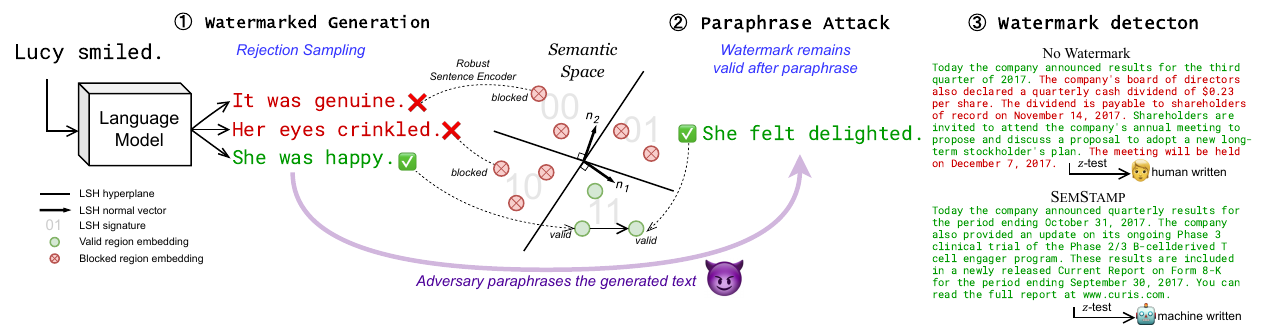}
    \caption{An overview of the proposed \method\ algorithm. \textbf{Left}: During generation, the watermark is injected by mapping candidate sentences into embeddings through a robust sentence encoder, dividing the semantic space through locality-sensitive hashing, and rejection sampling from the LM to generate sentences with valid region embeddings. \textbf{Right}: Detection is determined by the number of valid sentences in a candidate generation.}
    \figvsbottom
    \label{fig:teaser}
\end{figure*}

We propose \method, a \emph{semantic watermark algorithm} that is robust to sentence-level paraphrase attacks (\Sref{subsec:semantic_wm}). 
{Depicted in Figure \ref{fig:teaser}, our core intuition is that while paraphrasing alters the surface-form tokens, the sentence-level semantics are unchanged}. Thus, instead of partitioning the vocabulary, our watermark operates on the semantic space of sentence embeddings, partitioned by locality-sensitive hashing \citep[LSH;][]{indyk98lsh, charikar2002similarity}. We develop two key components---a sentence encoder trained with contrastive learning \citep[CL;][] {wieting-etal-2022-paraphrastic} and a margin-based constraint---to enhance paraphrastic robustness.

To stress-test the robustness of watermarking algorithms, we develop a novel attack method that minimizes bigram overlap during paraphrasing, and name it the bigram paraphrase attack (\Sref{subsec:bigram_attack}). 
Experimental results~(\Sref{sec:exp}) demonstrate that our proposed semantic watermark remains effective while token-level watermarks suffer significantly from the bigram attack. 

We summarize our main contributions as follows. First, we propose a sentence-level semantic watermark for LLMs and show that it is {robust to paraphrasing and more quality-preserving than a token-level watermark algorithm}. Second, we develop a novel attack method for watermarking algorithms, namely the bigram paraphrase attack, which can {effectively weaken token-level watermarking but only poses minor degradations to our semantic watermark}. Third, we fine-tune a paraphrase-robust sentence encoder with a contrastive learning objective and develop a rejection margin constraint to enhance the paraphrastic robustness of our semantic watermark algorithm.\footnote{Our code, model, and data will be released publicly.}

\secvsabove
\section{Approach}
\secvsbelow
\label{sec:approach}

\subsection{Preliminaries}
\label{subsec:prelim}
\subsecvs
\paragraph{Text Generation from Autoregressive LMs}
An autoregressive LM, denoted by $\LM$, models the conditional distribution of the next token over the vocabulary $V$. Given a token history $w_{1:t}=w_1,\dots,w_t$ where each token $w_i\in V$, the next token is generated by sampling $w_{t+1}\sim\LM(\cdot|w_{1:t})$. 
We introduce a sentence-level notation: $s^{(t+1)}\sim \LM(\cdot|s^{(1)}\dots s^{(t)})$ refers to the sampling of the next sentence given sentence history $s^{(1)}\dots s^{(t)}$. 

\paravs
\paragraph{Detecting Machine-Generated Text through Watermarking}


The goal of watermarked generation \citep[\textit{i.a.}]{Kuditipudi2023RobustDW,zhao2023provable} is to facilitate the detection of machine-generated text. 
A watermarked generation algorithm adds a statistical signal during the decoding stage of LLMs. The watermarked text is then provided to the user. At the detection stage, a piece of text is classified as machine-generated if the watermark is detected. 
Because malicious users could postprocess LLM-generated texts before detection, it is crucial that the watermark remains detectable under 
various text perturbations attacks, including text insertion, substitution, deletion, and paraphrasing. 


\paravs
\paragraph{Token-Level Watermarking and its Susceptibility to Paraphrase Attacks} 
\citet{kirchenbauer2023watermark} propose a watermark that is injected at the token level. At each time step of the generation, the vocabulary $V$ is pseudorandomly partitioned into a ``green list'' and a ``red list''. The random seed for partition is computed by a hash of the previously generated token. A globally fixed bias parameter $\delta>0$ is added to the logit of each green list token so that the LLM is induced to generate more green list tokens. The watermark is detected by conducting one proportion $z$-test (detailed in \Sref{app:z-score}) on the number of green list tokens in the generated text. 

Because of the token-level nature of the watermark algorithm, perturbing a token $w_t$ in a generated sequence $w_{1:T}$ through paraphrasing would change the green list for token $w_{t+1}$. As a result, a green token $w_{t+1}$ might be considered red, which undermines the detectability of the watermark \citep{krishna2023paraphrasing}. Moreover, because the watermark changes logits directly, it can degrade the quality of generated text \citep{Fu2023WatermarkingCT}. 

\begin{algorithm*}[t]
\small
\caption{\method{} text generation algorithm}
\label{alg:main}
\begin{algorithmic} 
\State \textbf{Input:} language model $\LM$, prompt $s^{(0)}$, number of sentences to generate $T$.

\State \textbf{Params:} sentence embedding model $\embd$ with embedding dimension $h$, maxout number $N_\text{max}$,
margin $m>0$, valid region ratio $\gamma\in(0,1)$, LSH dimension $d$, a large prime number $p$.
\State \textbf{Output:} generated sequence $s^{(1)}\dots s^{(T)}$.
\item[]

\Procedure{\method}{}
\State \textbf{init} $\textsc{Lsh}(\cdot)$, 
randomly initialize $d$ vectors
$n^{(1)}\dots n^{(d)}\in\mathbb{R}^h$, to create $2^d$ semantic subspaces.
\For{$t=1,2,\dots, T$}
 \begin{enumerate}[leftmargin=16.5mm,align=left]
\item Compute the LSH signature of the previously generated sentence, \textsc{Sig}($s^{(t-1)}$), and use $[\textsc{Sig}(s^{(t-1)})]_{10}\cdot p$ as the seed to randomly divide the space of signatures $\{0,1\}^d$ into a ``valid region set'' $G^{(t)}$ of size $\gamma\cdot 2^d$ and a ``blocked region set'' $R^{(t)}$ of size $(1-\gamma)\cdot 2^d$.
\item {\textbf{repeat} Sample a new sentence from LM,     
    
    \textbf{until} the signature of the new sentence is in the ``valid region set'', \textsc{Sig}($s^{(t)}$) $\in G^{(t)}$} and the margin requirement 
    
    \vspace{-0.5mm}\hspace{6.25mm} \textsc{Margin}($s^{(t)}, m$) is satisfied.
    

    \textbf{or} has repeated $N_\text{max}$ times

\item Append the selected sentence $s^{(t)}$ to context.
\end{enumerate}
\EndFor
\State \textbf{return} $s^{(1)}\dots s^{(T)}$
\EndProcedure
\end{algorithmic}
\end{algorithm*}

\begin{algorithm}[t]
\small
\caption{\method\ subroutines}
\label{alg:subroutine}
\begin{algorithmic}
\Function{Sig}{$s$} 
    \State $v \gets \embd(s)$~ // obtain embeddings of sentence $s$  
    \State $c \gets \textsc{Lsh}(v)$~ // obtain signature $c$ of the embedding  
    \State \textbf{return} $c$
\EndFunction
\item[]
\Function{Margin}{$s, m$} 
    \State $v \gets \embd(s)$~ // obtain embeddings of sentence $s$  

    \State $x\gets\min_{i=1,\dots,d}\{|\cos(v, n^{(i)})|\}$~ // {compute the minimum distance between $v$ and all LSH normal vectors $n^{(i)}$.}
    \State \textbf{return} \texttt{True} \textbf{If} $x\geq m$ \textbf{Else} \texttt{False}
\EndFunction
\end{algorithmic}
\end{algorithm}
\vspace{-2mm}

\paravs
\paragraph{Locality-Sensitive Hashing} We will use LSH \citep{indyk98lsh} to partition the semantic embedding space. It hashes similar inputs into similar signatures, thereby reducing the dimensionality and providing a similarity measure for a high-dimensional input space $\mathbb{R}^h$. Given an LSH dimension $d$, we adopt the cosine-preserving method from \citet{charikar2002similarity} which produces a $d$-bit binary signature through random hyperplane projections, and each hyperplane is represented by a random normal vector $n^{(i)}$ drawn from the $h$-dimensional Gaussian distribution.\footnote{Normal vector $n^{(i)}\in\mathbb{R}^h$ represents the hyperplane that is orthogonal to $n^{(i)}$ and passes through the origin.} 
The LSH signature for an embedding vector $v\in\mathbb{R}^h$ is then determined by the sign of the dot product between the candidate vector and the normal vectors: $\textsc{Lsh}_i:\mathbb{R}^h\mapsto\{0,1\}$ which gives the $i$-th digit signature, is defined by
$\textsc{Lsh}_i(v) = \mathbbm{1}\bigl( n^{(i)}\cdot v > 0\bigr)$\footnote{$\mathbbm{1}(\cdot)$ is the indicator function.}, and $\textsc{Lsh}(v)=[\textsc{Lsh}_1(v)||\dots||\textsc{Lsh}_d(v)]$ is the concatenation of all $d$ digits.


\subsection{\method: A Semantic Watermark with Paraphrastic Robustness}
\subsecvs
\label{subsec:semantic_wm}

We begin with a high-level overview of the \method\ (Alg.~\ref{alg:main}). Our approach is motivated by the intuition that paraphrasing alters the surface-form tokens but preserves sentence-level semantics. We apply the watermark at the sentence-level semantic space (instead of the token-level vocabulary) to preserve the watermark under token changes.
To do so, we use a semantic sentence encoder $\embd$ that produces vectors in $\mathbb{R}^h$. 
In practice, we fine-tune an off-the-shelf encoder with a contrastive objective \citep{wieting-etal-2022-paraphrastic} for paraphrastic robustness.

During the initialization of \method\ watermarked generation, we partition the space of sentence embeddings (produced by $\embd$) 
with the LSH introduced in \Sref{subsec:prelim}. Concretely, we initialize the $\textsc{Lsh}:\mathbb{R}^h\mapsto\{0,1\}^d$ function by sampling normal vectors $n^{(1)}\dots n^{(d)}$ to represent $d$  hyperplanes, and treat the space of LSH signatures $\{0,1\}^d$ as a natural partitioning of $\mathbb{R}^h$ into $2^d$ regions.

At each generation step, given a sentence history $s^{(0)}\dots s^{(t-1)}$, we first produce the LSH signature of the previously generated sentence $\textsc{Sig}(s^{(t-1)})$, where $\textsc{Sig}(\cdot)$ encodes and LSH-hashes the sentence, as defined in Alg.~\ref{alg:subroutine}. Next, we pseudorandomly divide the LSH partitions into a set of ``valid'' regions $G^{(t)}$ and a set of ``blocked'' regions $R^{(t)}$, where the masking is seeded by $\textsc{Sig}(s^{(t-1)})$.\footnote{\citet{kirchenbauer2023watermark} use ``green/red'' for vocabulary split. Instead, we adopt ``valid/blocked'' as the terminology for semantic region partition to be more accessible.} To produce the watermarked next sentence, we sample with rejection a new sentence $s^{(t)}$ from the LM until its embedding lies in the ``valid'' region in the semantic space.\footnote{We set a maxout number $N_\text{max}$ so that if there is still no valid sentence after sampling $N_\text{max}$ times, we choose the last sample as the next sentence.}

To detect the \method\ watermark, we conduct a one-proportion $z$-test on the number of valid-region sentences in the generated text. Since this detection is similar to \citet{kirchenbauer2023watermark}, we defer the details to \Sref{app:z-score}. 

Because a proper paraphrase should retain the meaning of the original sentence, we hypothesize that the LSH signature is likely to remain the same after paraphrasing (Figure \ref{fig:diffmargin} provides empirical results). Therefore, the valid region partition for the next sentence would not change, ensuring the watermark is still detectable after the paraphrase attack. Below we explain each core component of \method\ in detail.

\paravs
\paragraph{Paraphrase-Robust Sentence Encoder} 
A requirement for \method\ is a semantic encoder to map 
sentences into semantic embeddings. 
Our encoder is built upon Sentence-BERT \citep[SBERT;][]{reimers-gurevych-2019-sentence}, a fine-tuned siamese network trained to produce sentence embeddings whose cosine similarity mirror the semantic 
similarity of the STS benchmark \citep{cer-etal-2017-semeval}.

To enhance the encoder's robustness to paraphrase, we further fine-tune the SBERT model using contrastive learning \cite{wieting-etal-2022-paraphrastic}. For each sentence $s_i$ in a corpus, we first produce its paraphrase $t_i$ using an off-the-shelf paraphrasing model, Pegasus \citep{zhang2020pegasus}.\footnote{Link to 
\href{https://huggingface.co/tuner007/pegasus_paraphrase}{Pegasus paraphraser}.
} Next, we sample a random sentence $t'_i$ from the corpus that is not a paraphrase of $s_i$ to serve as the negative example. The objective promotes the original sentence to be more similar to the paraphrase than the negative example by a margin of $\delta>0$:
\begin{equation}
    \min_\theta \sum_i\max\Bigl\{\delta - f_\theta(s_i, t_i) + f_\theta(s_i, t'_i) , 0\Bigr\},
    \label{eq:clobj}
    \eqvs
\end{equation}
where $f_\theta$ is the cosine similarity between the embedded sentences, $f_\theta(s, t) = \cos\bigl(M_\theta(s),M_\theta(t) \bigr)$, and $M_\theta$ is the encoder model with parameter $\theta$. 

\paravs
\paragraph{Semantic Space Partitioning through LSH} During  the initialization of watermarked generation, normal vectors $n^{(1)}\dots n^{(d)}$ are randomly drawn from the $h$-dimensional Gaussian distribution to represent $d$ LSH hyperplanes in the semantic space  $\mathbb{R}^h$. The hyperplanes are fixed during generation and detection to serve as the basis for partitioning. As introduced in \Sref{subsec:prelim}, this induces a $d$-bit binary signature $\textsc{Lsh}(v)$ for a vector $v\in\mathbb{R}^h$. Consequently, we use each of the $2^d$ signatures $c\in\{0,1\}^d$ to represent a region in the semantic space consisting of points with signature $c$.

During the generation of a new sentence $s^{(t)}$, we apply a watermarking ``mask'' on the semantic space by pseudorandomly partitioning the space of signatures $\{0, 1\}^d$ into a valid region set $G^{(t)}$ of size $\gamma\cdot2^d$ and a blocked region set $R^{(t)}$ of size $(1-\gamma)\cdot2^d$, where $\gamma\in(0,1)$ determines the ratio of valid regions. The masking is seeded by the LSH signature of the last sentence $s^{(t-1)}$ and thus varies for each time-step $t$. Specifically, we convert the binary signature $\textsc{Sig}(s^{(t-1)})$ to decimal and use $[\textsc{Sig}(s^{(t-1)})]_{10}\times p$ to seed the randomization. Here $p$ is a large prime number
and $[.]_{10}$ an operator that casts binary numbers to decimal numbers. The condition for rejection sampling is that the LSH signature of the new sentence must fall into one of the valid regions, i.e., $\textsc{Lsh}(\embd(s^{(t)})\in G^{(t)}$.

\paravs
\paragraph{Margin-Based Constraint for 
Robustness} 
For the \method\ algorithm to be robust, the LSH signature of the sentences should remain the same under paraphrase attack. 
Empirically, we found the robustness from contrastive learning (Eq.~\ref{eq:clobj}) is not strong enough to preserve consistent LSH signature under paraphrasing. Therefore, we add an additional rejection sampling requirement that the sampled sentence $s^{(t)}$ must have the absolute value of cosine similarity with each normal vector $n^{(i)}$ larger than a margin $m>0$:
\begin{equation}
    \vspace{-1mm}
    \eqvs
    \min_{i=1,\dots,d}|\cos(n^{(i)}, v_t)| > m,
\end{equation}
where $v_t = \embd(s^{(t)})$ is the embedding of the candidate next sentence.\footnote{We discuss additional details on the condition for consistent LSH signature in \Sref{app:add_details}.}

\begin{figure}
    \begin{center}
    \vspace{-2mm}
    \includegraphics[width=0.32\textwidth]{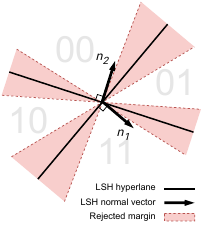}   
    \vspace{-4mm}
    \end{center}

  \caption{An illustration for margin-based rejection. Sentence embeddings at LSH hyperplane boundaries are rejected (highlighted in red).} 
  \vspace{-2mm}
  \label{fig:margin}
\end{figure}

Visually, this is akin to rejecting sentences whose embeddings lie near the boundaries of an LSH hyperplane. We illustrate this in Figure \ref{fig:margin}. In our experiments (\Sref{sec:exp}), we show that this margin-based rejection requirement can effectively increase the LSH signature robustness under paraphrasing.

\subsection{The Bigram Paraphrase Attack}
\label{subsec:bigram_attack}

We develop a strong ``bigram'' paraphrase attack with the following intuition.
Because existing token-level watermark algorithms hash the last generated token to determine the watermarking signature 
\citep{kirchenbauer2023watermark}, 
any choice of token at position $t$ would affect the watermark of position $t+1$. Therefore, we hypothesize that token-level watermarks might be especially sensitive to bigram (two adjacent tokens) perturbation.

Motivated by this intuition, we propose and explore the bigram paraphrase attack, a simple yet effective variant of the basic sentence-level paraphrase attack. Specifically, given a neural paraphrase model, we first decode a large number of top-raking sequences $s'_1\dots s'_k$ with beam search, obtaining $k$ paraphrase candidates. Next, we select the candidate that has the smallest bigram overlap with the original sentence. Moreover, to preserve the paraphrasing quality, we constrain the paraphrase attack with BERTScore  \citep{zhang2019bertscore} between paraphrases and original sentences:
\begin{align*}
    &s'= \argmin_{x\in\{s'_1,\dots,s'_k\}} \quad  \mathcal{B}(x, s), \\
    &\textrm{subject to} \quad \mathcal{S}(s'_1,s) - \mathcal{S}(x,s) \leq \Delta \cdot  \mathcal{S}(s'_1,s),
\end{align*}
where $s$ denotes the original sentence, $\mathcal{B}(x,s)$ is a simple counting of overlapped bigrams between sequences $x$ and $s$, $\mathcal{S}(x,s)$ denotes the BERTScore between sequence $x$ and $s$,
and $\Delta$ is the BERTScore threshold ratio. See Figure \ref{fig:textexamples} for an example in action.

\secvsabove
\section{Experiments}
\secvsbelow
\label{sec:exp}

\subsection{Experimental Setup}
\subsecvs

\paragraph{Datasets}

We conduct experiments to validate the detection robustness and quality of \method\ on the RealNews subset of the C4 dataset \citep{raffel2020exploring}, BookSum \citep{kryscinski2021booksum}, and Reddit-TIFU \citep{kim2019reddit-tifu}, a dataset with informal text style. We further analyze the detection results and generation quality on 1000 random samples. 

\paragraph{Metrics}
We use binary classification metrics: (1) area under the receiver operating characteristic curve (\textbf{\textit{AUC}}), and (2) the true positive rate when the false positive rate is 1\% or 5\% (\textbf{\textit{TP@1\%}}, \textbf{\textit{TP@5\%}}), i.e., the percentage of machine-generated text (the ``positive'' class in the classification setting) that is correctly detected when 1\% and 5\% of human texts (the ``negative'' class) are misclassified as machine-generated texts. A piece of text is classified as machine-generated when its $z$-score exceeds a threshold chosen based on a given false positive rate, which we explain in detail in \Sref{app:z-score}. Differing from KGW algorithm \citep{kirchenbauer2023watermark}, our algorithm treat sentences as the unit during $z$-score computation. 

To evaluate generation quality, we measure the perplexity (\textbf{\textit{PPL}}) with OPT-2.7B \citep{zhang2022opt}. Generation diversity is measured with trigram text entropy \citep{Zhang2018GeneratingIA} (\textbf{\textit{Ent-3}}), i.e., the entropy of the trigram frequency distribution of the generated text. We also evaluate generations with \textbf{\textit{Sem-Ent}} \citep{sement2022}, an automatic metric for semantic diversity. Following the setup in \citet{sement2022}, we use the last hidden states of OPT-2.7B models on generations as their semantic representation and perform $k$-means clustering.
Sem-Ent is the entropy of semantic cluster assignments of test generations.
We evaluate the quality of paraphrases using BERTScore \citep{zhang2019bertscore} between original generations and their paraphrases. 

\begin{table*}[t]
\centering
\footnotesize
\begin{tabular}{@{}lc | lll @{}}
\toprule
\multicolumn{2}{c}{} &
\multicolumn{3}{c}{{\texttt{RealNews | BookSum | Reddit-TIFU}}} 
\\
\cmidrule(lr){3-5}
\textit{\textbf{Paraphraser}} & \textit{\textbf{Algorithm}} & ~~~~~~~
\textit{\textbf{AUC}} $\uparrow$ & ~~~~~  
\textbf{\textit{TP@1\%}} $\uparrow$ & ~~~~~
\textbf{\textit{TP@5\%}} $\uparrow$ \\
\midrule 
&KGW &99.6 | 99.9 | 99.3 &98.4 | 99.4 | 97.5 &98.9 | 99.5 | 98.1\\ [-1ex] 
\raisebox{1.5ex}{No Paraphrase} &\shortmethod &99.2 | 99.7 | 99.7 &93.9 | 98.8 | 97.7 &97.1 | 99.1 | 98.2 \\[0.5ex]
\midrule\midrule
&KGW  &95.9 | 97.3 | 94.1 &82.1 | 89.7 | 87.2	&91.0 | 95.3 | 87.2 \\ [-1ex]
\raisebox{1.5ex}{Pegasus}  &\shortmethod &\textbf{97.8} | \textbf{99.2} | \textbf{98.4} &\textbf{83.7} | \textbf{90.1} | \textbf{92.8} &\textbf{92.0} | \textbf{96.8} | \textbf{95.4} \\[1ex]
&KGW  &92.1 | 96.5 | 91.7 &42.7 | 56.6	| 67.2 &72.9 | 85.3 | 67.6\\  [-1ex]
\raisebox{1.5ex}{Pegasus-bigram} &\shortmethod &\textbf{96.5} | \textbf{98.9} | \textbf{98.0}	&\textbf{76.7} | \textbf{86.8} | \textbf{89.0}	&\textbf{86.0} | \textbf{94.6} | \textbf{92.9} \\[1ex] \midrule
&KGW  &88.5	| 94.6 | 79.5 &31.5 | 42.0 | 22.8	&55.4 | 75.8 | 43.3\\[-1ex]
\raisebox{1.5ex}{Parrot} &\shortmethod &\textbf{93.3} | \textbf{97.5}  | \textbf{90.2} &\textbf{56.2} | \textbf{70.3} | \textbf{56.2} &\textbf{75.5} | \textbf{88.5} | \textbf{70.5} \\[1ex]
&KGW  &83.0 | 93.1	| 82.8 &15.0 | 39.9 | 27.6 &37.4 | 71.2 | 49.7\\[-1ex]
\raisebox{1.5ex}{Parrot-bigram} &\shortmethod &\textbf{93.1} | \textbf{97.5} | \textbf{93.9}	&\textbf{54.4} | \textbf{71.4} | \textbf{71.8} &\textbf{74.0} | \textbf{89.4} | \textbf{82.3} \\[1ex]
\midrule
&KGW  &82.8	| 87.6 | 84.1 &17.4 | 17.2 | 27.3	&46.7 | 52.1 | 50.9\\[-1ex]	
\raisebox{1.5ex}{GPT3.5} &\shortmethod &\textbf{83.3} | \textbf{91.8} | \textbf{87.7}	&\textbf{33.9} | \textbf{55.0} | \textbf{47.5} &\textbf{52.9} | \textbf{70.8} | \textbf{58.2} \\[1ex]
&KGW  &75.1	| 77.1 | 79.8 &5.9 \ \ | 4.4 \ \ | 19.3	&26.3 | 27.1 | 41.3\\[-1ex]
\raisebox{1.5ex}{GPT3.5-bigram} &\shortmethod{} &\textbf{82.2} | \textbf{90.5} | \textbf{87.4} 	&\textbf{31.3} | \textbf{47.4} | \textbf{43.8} &\textbf{48.7} | \textbf{63.6} | \textbf{55.9}\\[1ex]
\toprule 
\end{tabular}
\vspace{-3mm}
\caption{Detection results under different paraphraser settings. 
All numbers are in percentages. 
$\uparrow$ indicates higher values are preferred.
The numbers in parenthesis show the changes over our baseline.
\textbf{\method\ is more robust than KGW on multiple paraphrasers, datasets, and both the regular and bigram paraphrase attacks.}}
\label{tab:detection} 
\vspace{-2mm}
\end{table*}

\begin{table}
    \centering
    \footnotesize
    \begin{tabular}{rcccc}
        \toprule
        & \textbf{\textit{PPL}}$\downarrow$ & \textbf{\textit{Ent-3}}$\uparrow$ & \textbf{\textit{Sem-Ent}}$\uparrow$ \\ \midrule
        No watermark & 10.02 & 12.17 & 5.53 \\
        KGW & 12.17 & 12.10 & 5.47  \\
        \method & 10.20 & 12.16 & 5.51 \\ \bottomrule
    \end{tabular}
    \vspace{-2mm}
    \caption{Quality evaluation results. 
    $\uparrow$ and $\downarrow$ indicate the direction of preference (higher and lower). 
    \textbf{\method\ preserves the quality of generated text}.
    } 
    \vspace{-2mm}
    \label{tab:quality}
\end{table}

\paravs
\paragraph{Training, Generation, and Baselines}
For contrastive learning of SBERT, we paraphrase 8k paragraphs of the RealNews dataset \citep{raffel2020exploring} using the Pegasus paraphraser \citep{zhang2020pegasus} through beam search with 25 beams. We then fine-tune an SBERT model\footnote{sentence-transformers/all-mpnet-base-v1} with an embedding dimension $h=768$ on this subset for 3 epochs with a learning rate of $4 \times 10^{-5}$, using contrastive learning objective (Eq.~\ref{eq:clobj}). We set the contrastive learning margin $\delta = 0.8$ which is tuned from the dev set.

\begin{figure}[t]
    \centering
    \vspace{-3mm}
    \includegraphics[scale=0.3]{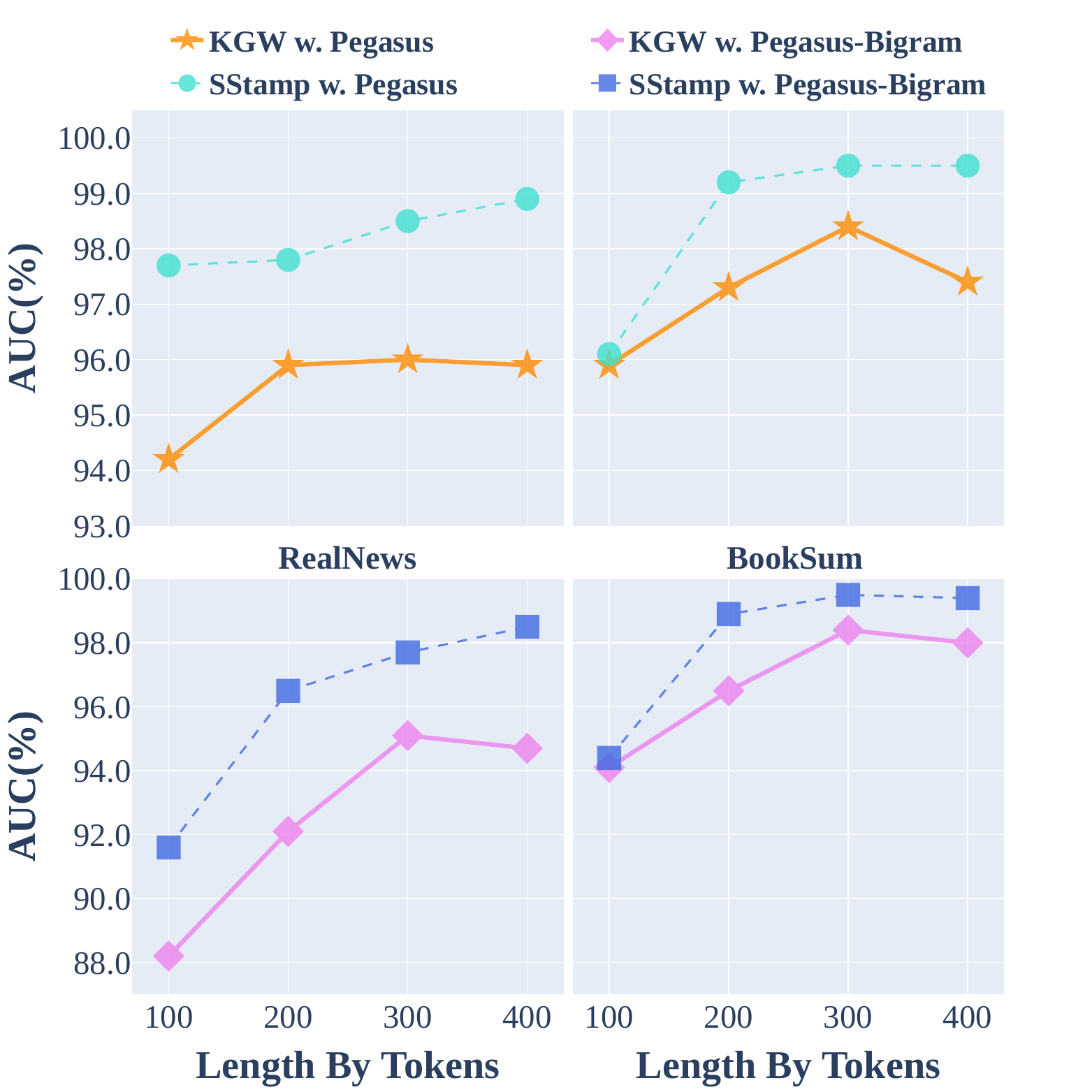}
    \vspace{-7mm}
    \caption{Detection results (AUC) under different generation lengths. \textbf{\method \ is more robust than KGW across length 100-400 tokens.}} 
    \vspace{-5mm}
    \label{fig:length}
\end{figure}

For watermarked generation, we use a fine-tuned version of OPT-1.3B \citep{zhang2022opt} as our base model to produce text with shorter length per sentence and conduct sampling at a temperature of 0.7 following \citet{kirchenbauer2023watermark} with a repetition penalty of 1.05. Setting 32 as the prompt length, we let 200 be our default generation length but also experiment on various different lengths (Fig.~\ref{fig:length}). To generate from \method, we sample at a LSH dimension $d=3$ with valid region ratio $\gamma = 0.25$ and rejection margin $m = 0.02$.  See \Sref{subsec:ablation} for the impact on hyperparameter choices. 

We choose the popular watermarking algorithm \citet[KGW;][]{kirchenbauer2023watermark} as our main baseline. In the paraphrase attack phase, we paraphrase generations by \method \ and KGW and compare their post-hoc detection rates after attacks. We also experiment with a distortion-free watermark by \citet[KTH;][]{Kuditipudi2023RobustDW} and \textsc{Unigram-Watermark} \citep{zhao2023provable}, but preliminary results show that KTH performs poorly compared to both KGW and \method \ against our paraphrase attacks for the AUC metric. \textsc{Unigram-Watermark} also demonstrates strong robustness against paraphrase, but it is vulnerable to being reverse-engineered (also see \Sref{app:add_exp_results}). 

\paravs
\paragraph{Paraphrase Attack}
For paraphrase attack experiments, watermarked generations are paraphrased sentence-by-sentence with the Pegasus paraphraser \citep{zhang2020pegasus}, the Parrot paraphrase used in \citet{sadasivan2023aigenerated}, and GPT-3.5-Turbo \citep{openai2022chatgpt}. We use beam search with 25 beams for both Pegasus and Parrot. For GPT-3.5-Turbo, we provide the sentences before the current sentence as the context and prompt the model to paraphrase via the OpenAI API.\footnote{\url{https://platform.openai.com/playground/}} A detailed description of prompts is included in \Sref{app:add_details}.

To implement the bigram paraphrase attack, we prompt the GPT-3.5-Turbo to return 10 paraphrases of the same sentence. For the Pegasus and Parrot paraphrasers, we select the candidate sentence with the least bigram overlap among the 25 beams from beam-search, subject to a BERTScore constraint of dropping no more than 10\% of the score from the first beam. For GPT-3.5-Turbo, the paraphrase sample with the highest BERTScore is treated as the first beam.

\subsecvs
\subsection{Results}
\secvsbelow
\label{exp:results}
\paragraph{Detection} Table \ref{tab:detection} shows detection results under different paraphrasers and the bigram attack at generation length 200. \textbf{\method\ is more robust to paraphrase attacks than KGW across the Pegasus, Parrot, and GPT-3.5-Turbo paraphrasers, as measured by AUC, TP@1\%, and TP@5\%}. 
Although we only fine-tune the SBERT model on data from the Pegasus paraphraser, \method \ algorithm generalizes its robustness to different paraphrasers (Parrot, GPT-3.5-Turbo) and works on texts from different domains. 

\textbf{The bigram paraphrase attack effectively weakens the token-level KGW algorithm while \method \ is relatively unaffected.} Pegasus bigram attack can lower KGW's AUC by 7.9\% and TP@5\% by 27.1\% on RealNews, whereas \method\ only decreases by 3.5\% and 13.2\%.
Furthermore, the BERTScore for bigram paraphrase does not change drastically compared to the regular paraphrases (Table \ref{tab:bertscore} in \Sref{app:add_exp_results}), showing that the bigram paraphrase attack still preserves paraphrase quality due to the BERTScore constraints we add. \citet{kirchenbauer2023reliability} propose several alternative hashing schemes to the KGW algorithm. We conduct paraphrase attack experiments on a recommended scheme named SelfHash, and do not find visible improvements to KGW, thus omitting the results for brevity. 

\paravs
\paragraph{Quality} Table \ref{tab:quality} compares quality metrics of non-watermarked generations with KGW and \method \ generations. 
\textbf{While KGW notably degrades perplexity due to the token-level noise added to logits, the perplexity of \method\ generation is on par with the base model without watermarking.} This confirms our hypothesis that the sentence-level nature of \method\ is less disruptive of token selections and preserves the generation quality.
Figure \ref{fig:textexamples} and \ref{fig:extratextexamples} provide qualitative examples of \method \ generations and the bigram paraphrase attack. Compared to non-watermarked generation, the sentences are equally coherent and contextually sensible. 
\textbf{\method\ also preserves token and semantic diversity of generation compared to non-watermarked generation and KGW generation}, as measured by the Ent-3 and Sem-Ent metrics, respectively.

\paravs
\paragraph{Generation Length} Figure \ref{fig:length} highlights that \method\ is robust to both regular and bigram paraphrase attacks across different generation lengths as measured by the number of tokens. \method\ has consistently higher AUC than KGW \citep{kirchenbauer2023watermark}.

\begin{figure}
    \vspace{-7mm}
    \centering 
    \includegraphics[scale=0.45]{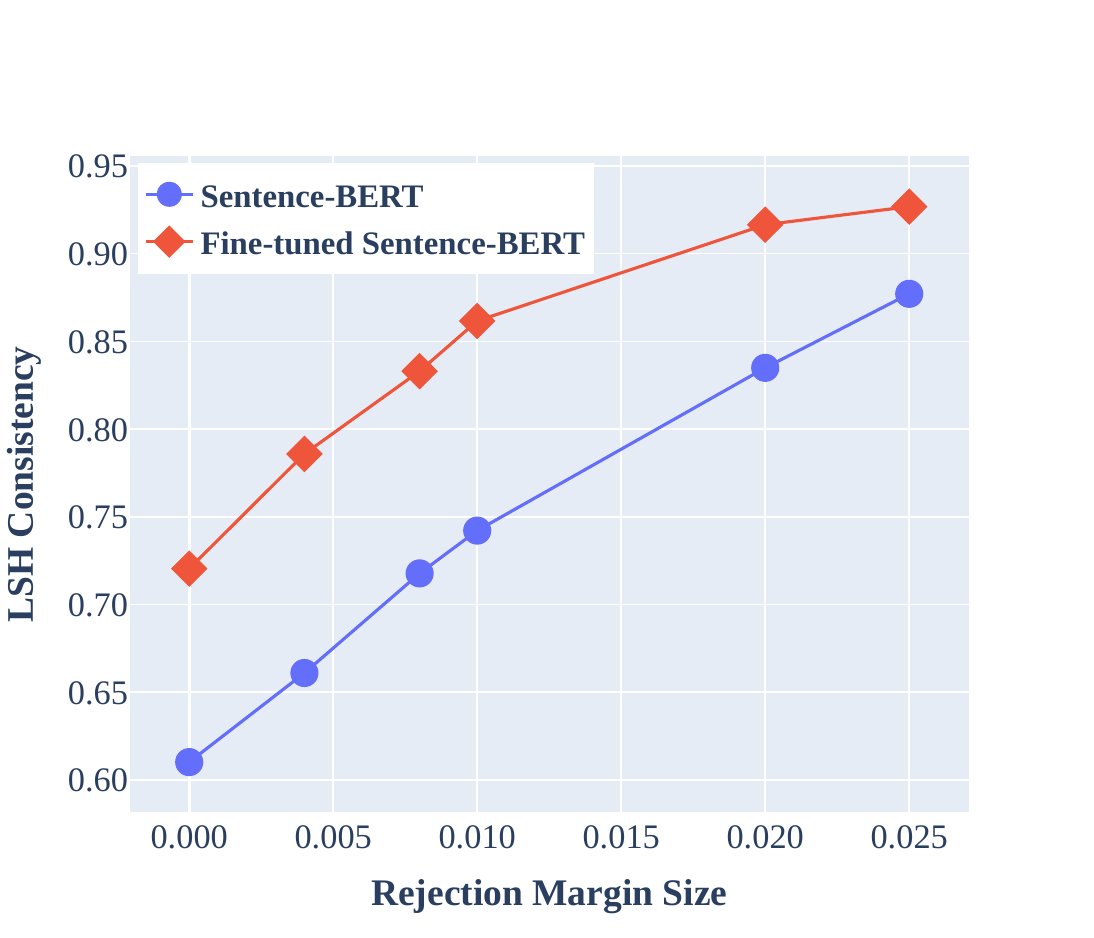}
    \vspace{-7mm}
    \caption{{Rejection margin and contrastive fine-tuning effectively improve LSH Consistency.}}
    \label{fig:diffmargin}
    \vspace{-5mm}
\end{figure}

\begin{figure*}[t]
    \centering
    \fbox{
    \begin{minipage}{38.5em}
    \footnotesize
    \textbf{Prompt: }
    Aylesbury’s new deputy area commander has been getting stuck in to his new role and says he’s determined to make himself ‘approachable’.
    \newline 
    \textbf{Non-Watermarked Generation: }
    Chief inspector Neil Kentish joined Thames Valley Police less than a week ago as deputy chief Constable. He’s already got his work cut out as the force tries to tackle crime in the post Brexit age. Here he shares his top tips for doing your job well. 1. Look the part – be respectful, look presentable, dress appropriately and be on time. Keep an eye out for the telly chaps who may be trying to use you for your money – you should never let this happen.
    \newline
    
    \textbf{\method : }
    Chief inspector Neil Kentish joined Thames Valley Police less than a month ago and has been thrust into the hot seat of the Aylesbury District Commander team. He was chosen from a wide range of candidates put forward by Deputy Chief Constable Peter Borsack, who said, “I am delighted to have appointed Neil as deputy chief officer of police in Aylesbury District. He is a real character and I believe he will be a great addition to the team."
    \newline
    \textbf{Pegasus Paraphrase: }
    Neil Kentish was the chief inspector of Thames Valley Police. "I was thrust into the hot seat of the district commander team a month ago." He was chosen from a wide range of candidates put forward by Peter Borsack who said: "I am delighted to have appointed Neil as deputy chief officer of police in Aylesbury District. I think he will be a great addition to the team."
    \newline
    \textbf{Pegasus Bigram Paraphrase: }
    Neil Kentish was the chief inspector of Thames Valley Police. He was put into the hot seat of the district commander team a month ago. Neil was chosen from a wide range of candidates put forward by Peter Borsack, who said he was delighted to have appointed Neil as deputy chief officer of police. "I think he will be a good addition to the team. He will bring a good level of leadership and management skills to the community."
    \end{minipage}
    } 
    \caption{Generation Examples. Paraphrase examples are based on \method \ generations. Additional examples are presented in Figure \ref{fig:extratextexamples} in the Appendix. \textbf{\method\ generations are equally coherent and contextually sensible compared to non-watermarked generations.}}
    \label{fig:textexamples}
    \vspace{-3mm}
\end{figure*}



\label{subsec:ablation}
\paravs
\paragraph{Analysis}
Figure \ref{fig:diffmargin} shows that increasing margin size $m$ will increase the consistency of LSH signatures (\textit{LSH consistency}), the ratio of sentences that remain in the same valid region after being paraphrased. A higher rejection margin will ensure the sampled generations are further away from the region boundary, thus less likely to shift to a different region after paraphrasing. However, a larger margin will result in a slower generation speed, and we find $m=0.02$ works well empirically.

We also compare the LSH consistency before and after fine-tuning SBERT with contrastive learning in Figure \ref{fig:diffmargin}. Fine-tuning the encoder on Pegasus-paraphrased data improves the LSH consistency across different margins.

Applying the masking of semantic space partitions and the rejection margin, \method\ makes a trade-off between watermark detection accuracy and generation speed. For our current hyperparameter setting, 13.8 sentences are needed on average to sample one valid sentence. As we explain in the Limitations and Discussion section, this limitation can be mitigated if we conduct batched sampling of next sentences.


\secvsabove
\section{Related Work}
\secvsbelow

Machine-generated text detection, aiming at distinguishing LLM-generated texts from human-written ones, can be categorized into proactive and post-hoc methods. Our focus, watermarked generation, belongs to the first category.

\paravs
\paragraph{Watermarked Generation}
Early approaches to watermarking include text-meaning representation tree for information hiding \citep{Atallah2002watermark}, and a watermarking scheme for machine translation using an output selector that considers hashing operation \citep{venugopal2011watermark}. Watermarked language generation, specifically pertinent to LLMs, is a renewed trend of proactive machine-generated text detection. The scheme works by adding signatures imperceptible to humans during decoding time to enable stable detection at a later time.
\citet{kirchenbauer2023watermark} propose a watermarking algorithm by adding token-level bias (reviewed in \Sref{sec:approach}). \citet{Kuditipudi2023RobustDW} proposes a distortion-free watermark that preserves the original distribution of LM during watermarking. 
\citet{Yoo2023RobustMN} embeds multi-bit information into watermark and enhances performance against corruption through a robust infilling model. 
They inject the watermark via word replacement after initial generation, which is incorporated into one-stage watermarked generation by \citet{Wang2023TowardsCT}. 
\citet{Christ2023UndetectableWF} propose a watermarking scheme that is computationally undetectable without the secret key in theory.

Importantly, these existing works employ a token-level design and focus on span-level corruption such as editing and cropping, which renders the watermarks susceptible to paraphrase attacks.

More related to our focus on paraphrase attack, \citet{krishna2023paraphrasing} propose a retrieval-based method that requires saving all previously-generated sequences, and \citet{kirchenbauer2023reliability} empirically shows that \citet{kirchenbauer2023watermark} is more robust under longer generation length. Contemporary to our work, \citet{zhao2023provable} improves robustness via a cryptographic-free watermark without hashing previous tokens, which is more robust to editing and paraphrasing attacks.
To the best of our knowledge, our work is the first sentence-level semantic watermark algorithm targeted against paraphrase attacks.

\paravs
\paragraph{Post-Hoc Detection of Machine-Generated Text}  In post-hoc methods, applying binary classification models is the most straightforward approach \citep{zellers2019neuralfakenews, jawahar-etal-2020-automatic, Liu2022CoCoCM, Mireshghallah2023SmallerLM, pu-etal-2023-zero}. These methods are applicable to black-box generators but need sufficiently large corpus for fine-tuning. Another type of post-hoc detection is based on statistical patterns within generation, including token likelihood \citep{gehrmann2019gltr}, rank \citep{Solaiman2019ReleaseSA}, entropy \citep{ippolito-etal-2020-automatic}, and likelihood gap at perturbation \citep{Mitchell2023DetectGPTZM, su2023detectllm}. These methods have better interpretability but are reliable only with white-box access to generators. \citet{sadasivan2023aigenerated} question the theoretical reliability of detection while \citet{chakraborty2023possibilities} support detection is achievable.

We defer further related works on LSH, watermarking for copyright, and contrastive learning to \Sref{app:related_work} due to space reasons.

\secvsabove
\section{Conclusion}
\secvsbelow

We introduce \method, a novel sentence-level semantic watermark for LLMs. The watermark is injected by mapping candidate sentences into embeddings with a paraphrase-robust encoder, partitioning the semantic space through LSH, and rejection sampling to generation sentences with valid region embeddings. Empirical results show that \method\ is not only robust to paraphrase attacks but also more quality-preserving than a token-level baseline watermark algorithm. We also propose a bigram paraphrase attack which effectively weakens the token-level watermark while only causing minor performance deterioration to \method. We hope \method\ can serve as an effective tool for regulating the proliferation of machine-generated texts.

\secvsabove
\section*{Limitations and Discussion}
\secvsbelow
\label{sec:discuss}

\paragraph{Robustness to Stronger Attacks} 
Since \method\ operates on the sentence level, it is not robust against attacks on the inter-sentence level. For example, a recently proposed paraphraser Dipper \citep{krishna2023paraphrasing} includes sentence reordering. Our algorithm is also less effective when the machine text is embedded in a relatively large portion of human text. We leave the exploration of stronger attacks to future work.

\paravs
\paragraph{Semantic Constraint from LSH} 
While the LSH partitioning divides the full semantic space into sub-regions, enforcing the ``valid region'' requirement during generation may potentially reduce the generation flexibility. Interestingly, we use a small LSH dimension ($d=3$) and we do not observe a visible quality degradation. A potential explanation is that with a smaller LSH dimension, the valid partition also becomes larger, which does not impose a strong semantic constraint and provides enough diversity for generations, as we found in our experiments (\Sref{exp:results}).

\paravs
\paragraph{Speed} 
Due to the nature of rejection sampling, text generation with \method\ is slower than non-watermarked generation by a factor of 20.9 with LSH dimension $d=3$ and margin $m=0.02$ (\Sref{exp:results}), and by a factor of 5.26 when $d=3$ and $m=0$ (Table \ref{tab:diffdim}). However, since candidate sentences for rejection sampling have the same LM context, it is possible to conduct batch sampling of candidate next sentences, which speeds up watermarked generation while increasing the memory overhead. We see the additional computation cost for \method\ as a cost for robustness: adding the watermark on the semantic space trades-off speed for better detection accuracy under paraphrase attacks. A potential mitigation to speed up is through parallel decoding multiple sentences across multiple GPUs, which makes our algorithm a viable option for cloud servers with ample computing. For future work, it would be exciting to make this procedure more efficient with techniques from controlled generation \citep{weir-etal-20-cod3s, yang-klein-2021-fudge, keskarCTRL2019}.

\paravs
\paragraph{Reverse Engineering} Since our sentence encoder and LSH hyperplanes are not public, it is not straightforward for a curious attacker to reverse engineer the configurations and we leave it for future work to explore. 
The difficulty of reverse engineering can also be increased by using a larger LSH dimension, while the watermark could be less robust to paraphrase attack.

\paravs
\paragraph{Bigram Paraphrase Attack Control}
We control the ``intensity'' degree of bigram paraphrase attack by constraining the paraphrase candidate selection with a BERTScore constraint. Removing the constraint will more forcefully lower AUROC at the expense of paraphrase quality.

Finally, due to lack of space we defer discussions on ethical impacts to \Sref{app:ethical}.

\secvsabove
\section*{Acknowledgement}
This research is supported in part by the Office of the Director of National Intelligence (ODNI), Intelligence Advanced Research Projects Activity (IARPA), via the HIATUS Program contract \#2022-22072200004.
This material is also funded by the DARPA Grant under Contract No.~HR001120C0124.
We also gratefully acknowledge support from NSF CAREER Grant No.~IIS2142739, NSF Grants No.~IIS2125201, IIS2203097, and the Alfred P.~Sloan Foundation Fellowship.
The views and conclusions contained herein are those of the authors and should not be interpreted as necessarily representing the official policies, either expressed or implied, of ODNI, IARPA, or the U.S. Government. The U.S.~Government is authorized to reproduce and distribute reprints for governmental purposes notwithstanding any copyright annotation therein.

\bibliography{custom,anthology}

\clearpage
\appendix

\section*{Supplemental Materials}

\section{Additional Related Works}
\label{app:related_work}
\paragraph{Locality-Sensitive Hashing in NLP}
The application of locality-sensitive hashing \citep{indyk98lsh, charikar2002similarity} in NLP dates back to \citet{ravichandran-etal-2005-randomized}, where LSH is used for high-speed noun clustering. \citet{van-durme-lall-2010-online} show that the LSH method of \citet{charikar2002similarity} can enable fast approximated online computation of cosine similarity. \citet{guu-etal-2018-generating} use LSH to efficiently compute lexically similar sentences in a prototype-then-edit sentence generation model. Closely related to our work, \citet{weir-etal-2020-cod3s} generate semantically diverse sentences by conditioning a sequence-to-sequence model on the LSH signature of sentence embeddings.

\paragraph{Watermarked Natural Language Data for Copyright}
Watermarked generation can be further applied for data copyright protection. 
\citet{gu2022watermarking} embed backdoor trigger words as black-box watermarks into LLMs.
\citet{Liu2023WatermarkingTD} propose a novel watermark via backdoor-based membership inference, where backdoor watermarked texts poison unauthorized training models.
\citet{Yao2023PromptCAREPC} focus on protecting the copyright of prompts through inserting the secret key into the prompt optimization stage. These works mainly apply watermark techniques for data copyright protections , whereas our work focuses on exploring the robustness of watermark against paraphrasing.





\paragraph{Contrastive Learning in NLP}
Contrastive learning \citep{1640964} aims at improving the distinguishability of representation by pulling over positive pairs and pushing off negative pairs. 
In the NLP domain, contrastive learning can be applied to sentence embedding \citep{logeswaran2018efficient}, and further used in downstream tasks like natural language inference \citep{Li2022PairLevelSC}, understanding \citep{Fang2020CERTCS}, reasoning \citep{Klein2020ContrastiveSL}, classification \citep{Choi2022C2LCC} etc. \citet{logeswaran2018efficient} apply unsupervised contrastive learning between current sentence candidates and context sentences to effectively learn sentence representation. 
\citet{gao2021simcse} further apply supervised contrastive learning in sentence embedding by using annotated pairs from natural language
inference.
\citet{Kim2021SelfGuidedCL} propose a self-guided contrastive learning between embeddings from a fixed model and a fine-tuned model.

\section{Watermark Detection}
\label{app:z-score}
\citet{kirchenbauer2023watermark} proposes using a one-proportion $z$-test on the number of green list tokens to detect watermarks, assuming the following null hypothesis:
\begin{align*}
    H_0: &~\textit{The text is not generated (or written)} \\
    &~ \textit{knowing a watermarking green list rule.}
\end{align*}
The null hypothesis is rejected when the $z$-score computed based on the number of green tokens in a piece of text $T$ exceeds a given threshold $M$:
\begin{equation}
\label{eq:z-score}
    z_{\text{KGW}} = \frac{N_G - \gamma N_T}{\sqrt{\gamma(1-\gamma)N_T}},
\end{equation} where $N_G$ denotes the number of green tokens, $N_T$ refers to the total number of tokens contained in the given piece of text $T$, and $\gamma$ is a chosen ratio of green tokens. 
During detection time, the number of green tokens in each piece of text will be counted. According to Eq. \ref{eq:z-score}, a higher ratio of detected green tokens means a higher $z$-score, determining with more confidence that the text is machine-generated.

We adapt this one proportion $z$-test to \method, modifying the null hypothesis and using sentence as our basic unit:
\begin{align*}
    &H_0:\\
    &~\textit{The text is not generated (or written) knowing}\\ 
    &~\textit{a rule of valid and blocked partitions in the} \\
    &~\textit{semantic space.}
\end{align*}
\begin{equation}
    z_{\method} = \frac{S_V - \gamma S_T}{\sqrt{\gamma(1-\gamma)S_T}},
\end{equation}
where $S_V$ refers to the number of valid \textit{sentences}, $\gamma$ is the ratio of valid sentences out of the total number of sentences $S_T$ in a piece of text $T$. 

During detection time, we first break a piece of texts into individual sentences and detect the number of valid sentences $S_V$ to calculate $z_{\method}$. We detect a machine-generated text when $z_{\method} > M_{r}$, where $M_{r}$ is located according to a given false positive rate $r$: We define machine-generated as the positive class in classical classification setting and non-machine-generated as the negative class. We iterate through a range of possible $m \in [0,6.0]$ until there is a $M_r=m$ such that $r$ percentage of human (negative-class) texts is misclassified as machine-generated. For example, we let $r=0.05$ for the \textbf{\textit{TP@5\%}} metric in Table \ref{tab:detection}.

\begin{figure*}
    \centering
    \fbox{
    \begin{minipage}{38.5em}
    \footnotesize
    \textbf{Prompt: }
    NEW DELHI: Over fifteen years after the Concorde — the world’s first and only supersonic aircraft to be used commercially — retired, US aerospace major Boeing has announced it is working on its successor.
    \newline 
    \textbf{Non-Watermarked Generation: }
    The first commercial supersonic jet was developed in the late 1960s by British Aerospace. However, the technology was never adopted by the industry. Boeing aims to change that. It has set up a team of engineers who will work on developing the next Supersonic Jet. 
    \newline
    \textbf{Baseline Watermark:}
    The company has called it the X-35. But it’s not yet clear when it might begin taking orders for production. The project is named after Jason Xtra, who first proposed it in 1997 – after the US Air Force expressed interest in it. However, Boeing hasn’t announced any orders for the X-35 yet.
    \newline\newline
    \textbf{\method :}
     The company said it was committed to developing the space elevator and had launched a concept for a space elevator in 2003. Boeing’s chief financial officer, Robert Lach Jr, said the company would spend about \$2 billion over the next five years on what is called the Space Elevator Initiative. Boeing estimated that an elevator would cost between \$8 billion and \$10 billion to build, depending on the design. 
    \newline\newline
    \textbf{Pegasus Paraphrase: }
   The company launched a concept for a space elevator in 2003 and said it was committed to developing the space elevator. Boeing will spend \$2 billion over the next five years on the Space Elevator Initiative, according to Robert Lach Jr., the company's chief financial officer. Depending on the design, an elevator could cost between \$8 billion and \$10 billion.
    \newline
    \textbf{Pegasus Bigram Paraphrase: }
    In 2003 the company launched a concept for a space elevator. The company will spend \$2 billion over the next five years on the Space Elevator Initiative. Depending on the design, an elevator could cost as much as \$10 billion.
    \end{minipage}
    } 
    \caption{Additional Generation Examples. Non-Watermarked refers to the original model without adding any watermark. Baseline Watermark refers to \cite{kirchenbauer2023watermark}. Paraphrase examples are based on \method \ generations.}
\label{fig:extratextexamples}
\end{figure*}

\section{Effect of LSH dimension $d$}
\label{app:lshdim}
\begin{table}[t]
\centering
\footnotesize
\begin{adjustbox}{width=\columnwidth,center}
\begin{tabular}{ccc} 
\toprule
{LSH Dim ($d$)} & {Average \# of Sentences Sampled $\downarrow$} & {LSH Consistency $\uparrow$ }\\
\midrule 
3 &5.26 &\textbf{.720} \\
4 &4.53	&.666	\\
8 &4.26 &.508 \\
16 &\textbf{4.14} &.335 \\
\bottomrule
\end{tabular}
\end{adjustbox}
\caption{Effects of Increasing LSH Dimensions at margin $m = 0.0$. The sampling rate is the average number of sentences sampled to produce one valid (watermarked) sentence.} 
\label{tab:diffdim}
\end{table}

In Table \ref{tab:diffdim}, we discover that fewer LSH dimensions will make a sentence more likely to stay in the same region after being paraphrased. We define LSH Consistency as the ratio of paraphrased sentences that have the same LSH signature as the original sentence over the total number of paraphrased sentences. A higher consistency ratio indicates better robustness.

Geometrically, when the LSH dimension is lower, there are fewer partitioned semantic regions, each having a larger space. A paraphrase will have a similar representation with its source sentence in the semantic space, which will be more likely to remain in the same semantic region if each region is larger.

On the other hand, lowering the number of LSH dimensions will also slightly increase the average number of sentences sampled to produce one valid sentence (Average Number of Sentences Sampled). We ultimately decide on a minor sacrifice in speed for the gain of accuracy and choose $d=3$. We choose $\gamma = 0.25$ following \citet{kirchenbauer2023watermark}, where the authors show that larger green-list ratios will lower the $z$-score.

\section{Additional Experimental Results}
\label{app:add_exp_results}
We include additional experimental results on paraphrase quality, i.e., the BERTScore between original and paraphrased generations under different settings, in Table \ref{tab:bertscore}.

We provide paraphrased detection results of the KTH algorithm \citep{Kuditipudi2023RobustDW} in Table \ref{tab:detection_with_kth} and \textsc{Unigram-Watermark} \citep{zhao2023provable} in Table \ref{tab:detection_with_uw}. We find that the KTH watermark performs poorly against KGW and \method. 

Although \textsc{Unigram-Watermark} enjoys strong robustness against paraphrasing attacks, it has the crucial limitation of being readily hacked by an adversary. Since Unigram-Watermark can be understood as a variant of KGW \citep{kirchenbauer2023watermark} but with only one fixed greenlist initialized at the onset of generation. An adversary can reverse-engineer this greenlist by brute-force submissions to the detection API of $|V|$ times, where each submission is repetition of a token drawn with out replacement from the vocabulary $V$ of the tokenizer. Therefore, upon each submission to the detection API, the adversary will be able to determine if the submitted token is in the greenlist. After $|V|$ times of submission, the entire greenlist can be reverse-engineered. On the other hand, such hacks are not applicable to \method, since \method\ does not fix the list of valid regions and blocked regions during generation, and the adversary needs to have access to the private sentence embedder and LSH hyperplanes to hack \method. 

In summary, despite having strong robustness against various paraphrase attacks, \textsc{Unigram-Watermark} has a notable vulnerability that may limit its applicability in high-stake domains where adversaries can conduct reverse-engineering.

\paragraph{Computing Infrastruture and Budget}
We run sampling and paraphrase attack jobs on 8 A40 GPUs, taking up a total of around 100 GPU hours.

\begin{table*}[]
\small
\centering
\begin{tabular}{lcccccc}
\toprule
 & \multicolumn{3}{c}{\texttt{RealNews}} & \multicolumn{3}{c}{\texttt{BookSum}} \\ \cmidrule(l){2-7} 
\multicolumn{1}{l|}{\textbf{\textit{Algorithm}}$\downarrow$~\textbf{\textit{Paraphraser}}$\rightarrow$} & Pegasus & Parrot & \multicolumn{1}{c|}{GPT3.5} & Pegasus & Parrot & GPT3.5 \\ \midrule
\multicolumn{1}{c|}{KGW} & 71.0 / 66.6 & 57.1 / 58.4 & \multicolumn{1}{c|}{54.8 / 53.3} & 71.8 / 69.3 & 62.0 / 61.8 & 60.3 / 56.7 \\
\multicolumn{1}{c|}{\shortmethod} & 72.2 / 69.7 & 57.2 / 57.4 & \multicolumn{1}{c|}{55.1 / 53.8} & 72.7 / 70.2 & 62.9 / 62.4 & 61.8 / 58.4 \\ \bottomrule
\end{tabular}
\caption{BERTScore between original and paraphrased generations under different settings. All numbers are in percentages. The first number in each entry is under vanilla paraphrase attack while the second number is under the bigram paraphrase attack. \textbf{Bigram paraphrase attack poses only minor degradation on semantic similarity with original sentence compared to vanilla paraphrase attack.}}
\label{tab:bertscore}
\end{table*}

\begin{table*}[]
\small
\centering
\begin{tabular}{lcccc}
\toprule
 & \multicolumn{2}{c}{\texttt{RealNews}} & \multicolumn{2}{c}{\texttt{BookSum}} \\ \cmidrule(l){2-5} 
\multicolumn{1}{l|}{\textbf{\textit{Algorithm}}$\downarrow$~\textbf{\textit{Paraphraser}}$\rightarrow$} & Pegasus & Parrot & Pegasus & Parrot  \\ \midrule
\multicolumn{1}{c|}{KGW} & 95.9 / 92.1 & 88.5 / 83.0 & 97.3 / 96.5 & 94.6 / 93.1  \\
\multicolumn{1}{c|}{\shortmethod} & 97.8 / 96.5 & 93.3 / 93.1 & 99.2 / 98.9 & 97.5 / 97.5 \\
\multicolumn{1}{c|}{\textsc{Unigram-Watermark}} & \textbf{99.1} / \textbf{98.4} & \textbf{98.9} / \textbf{98.7} & \textbf{99.4} / \textbf{99.7} & \textbf{99.5} / \textbf{99.6}\\ \bottomrule
\end{tabular}
\caption{Paraphrased detection results of \textsc{Unigram-Watermark}. We find that \textsc{Unigram-Watermark} demonstrates strong robustness against paraphrase attacks, but has the vulnerability to being reverse-engineered, which we discuss in \Sref{app:add_exp_results}.}
\label{tab:detection_with_uw}
\end{table*}

\begin{table}[t]
\centering
\footnotesize
\begin{tabular}{c | ccc }
\toprule
\multicolumn{1}{c}{} &
\multicolumn{3}{c}{{\texttt{BookSum}}}
\\
\cmidrule(lr){2-4}
\textit{\textbf{Algorithm}} &
\textit{\textbf{AUC}} $\uparrow$ &  
\textbf{\textit{TP@1\%}} $\uparrow$ &
\textbf{\textit{TP@5\%}} $\uparrow$
\\ 
\midrule 

KGW  &95.9 &82.1	&91.0\\
KTH &51.7 &5.0 &5.8 \\
\method &\textbf{97.8} &\textbf{83.7} &\textbf{92.0} \\
\toprule 
\end{tabular}
\caption{Paraphrased detection results on the BookSum dataset. The paraphraser used is Pegasus. We find that the KTH watermark performs poorly against KGW and \method.}
\label{tab:detection_with_kth} 
\vspace{-2mm}
\end{table}

\section{Additional Details}
\label{app:add_details}
\paragraph{Condition for consistent LSH signature}
For robustness, the \method\ algorithm would need the LSH signature of the paraphrased sentence to be unchanged from the signature of the original sentence. This requires that for each LSH digit $i$, the sign of the dot product between the embedded sentence and the normal vector $n^{(i)}$ should not change before and after paraphrasing:
\begin{equation}
    \begin{aligned}
        \mathbbm{1}\bigl(n^{(i)}\cdot v_\text{orig} >0\bigr) = \mathbbm{1}\bigl(n^{(i)}\cdot v_\text{para}>0\bigr),& \\
        \forall i\in\{1\dots d\},&
    \end{aligned}
\end{equation}
where $v_\text{orig}=\embd(s^{(t)})$ and $v_\text{para}=\embd(G(s^{(t)}))$ are the embeddings for the original and paraphrased sentences, respectively, and $G$ is the paraphraser.

\paragraph{Cosine Similarity} In \Sref{subsec:semantic_wm}, we slightly abuse the notation and use $\cos(\vx,\vy)$ to denote the \textit{cosine similarity} between two vectors $\vx$ and $\vy$. That is,
\begin{equation}
    \cos(\vx,\vy) = \frac{\vx\cdot\vy}{|\vx||\vy|}.
\end{equation}

\paragraph{Sentence Delimitation}
\label{app:sentence_delim}
During generation time, a full candidate next sentence is considered generated if the language model has generated a new delimiter punctuation, i.e., a comma, period, question mark, or exclamation mark.

\paragraph{Data Preprocessing} We separate the data points, which are paragraphs of news (RealNews) and book summaries (BookSum), into sentences using \texttt{nltk.sent\_tokenize}. Additionally, we add a period mark to every sentence that does not end in a comma, period, question mark, or exclamation mark.

\paragraph{Prompt for GPT-3.5-Turbo Paraphrase}

To use GPT-3.5-Turbo as a paraphraser, we provide the following prompt: 
\begin{center}
    \texttt{Previous context: \{context\} $\backslash$n \\ Current sentence to paraphrase: \{sent\}}
\end{center}

We define \texttt{sent} to be the target sentence to be paraphrased, and \texttt{context} as the list of sentences before the target sentence.

For the bigram paraphrase attack, we provide the following prompt:
\begin{center}
    \texttt{Previous context: \{context\} $\backslash$n Paraphrase in \{num-beams\} different ways and return a numbered list : \{sent\}}
\end{center}
where \texttt{num-beams} specifies the number of candidate sentences. A higher \texttt{num-beams} will strengthen the bigram paraphrase attack but also at the cost of more computational resources.

\section{Ethical Impacts}
\label{app:ethical}


As language models become increasingly capable of generating realistic texts, the risk of misusing language model generations, such as spreading misinformation, practicing plagiarism, and violating copyrights, has become imminent. Furthermore, on a fundamental level, the inability to distinguish humans from machines poses threats to establishing the basic level of mutual understanding and trust that bonds society. Robust detection of machine-generated text is crucial for preventing the misuse of large language models by properly attributing the source of online texts. Although current LLMs are often exposed to users as API endpoints, malicious users can still postprocess and paraphrase the API-generated response to escape the injected watermark. This motivates us to study watermark robustness against paraphrasing in this work. We hope that the proposed \method\ algorithm can mitigate the risk of LLM misuse by providing a reliable method to counter paraphrasing attacks on watermarked generations.
\end{document}